\documentclass[11pt,a4paper]{article}
\usepackage[hyperref]{acl2020}
\usepackage{times}
\usepackage{latexsym}

\usepackage{microtype}

\usepackage{biblatex} %Imports biblatex package
\title{Introducing Semantics into Speech Encoders}

\author{
Derek Xu\textsuperscript{1}, 
Shuyan Dong\textsuperscript{2},
Changhan Wang\textsuperscript{2}\Thanks{Equal Contribution},
Suyoun Kim\textsuperscript{2}\footnotemark[1],
Zhaojiang Lin\textsuperscript{2}\footnotemark[1],\\
Akshat Shrivastava\textsuperscript{2},
Shang-Wen Li\textsuperscript{2},
Liang-Hsuan Tseng\textsuperscript{3},
Alexei Baevski\textsuperscript{2},\\
Guan-Ting Lin\textsuperscript{3},
Hung-yi Lee\textsuperscript{3},
Yizhou Sun\textsuperscript{1},
Wei Wang\textsuperscript{1}
\\ \\
\textit{\textsuperscript{1}University of California, Los Angeles, \textsuperscript{2}Meta AI, \textsuperscript{3}National Taiwan University} \\ \\
\texttt{\small \textsuperscript{1}\{derekqxu, yzsun, weiwang\}@cs.ucla.edu} \\
\texttt{\small \textsuperscript{2}\{asyd,changhan,suyounkim,zhaojiang,akshats,shangwel,abaevski\}@meta.com} \\
\texttt{\small \textsuperscript{3}\{r11921067,r10942104,hungyilee\}@ntu.edu.tw} \\
}

\date{}

\usepackage{graphicx}
\usepackage{caption}
\usepackage{subcaption}
\usepackage{multirow}
\usepackage{amsmath}
\usepackage{amssymb}
\usepackage{graphicx}
\usepackage{xspace}
\usepackage{hyperref}

\newcommand{\derek}[1]{\ignorespaces}
\newcommand{\annie}[1]{\ignorespaces}

\newcommand{\wtvo}{\textsc{w2v}\xspace}
\newcommand{\wtv}{\textsc{w2v2}\xspace}
\newcommand{\wtvasr}{\textsc{w2v2-ASR}\xspace}
\newcommand{\wtvl}{\textsc{w2v2l14}\xspace}
\newcommand{\hubert}{\textsc{Hubert}\xspace}
\newcommand{\wtvuo}{\textsc{w2v2-U}\xspace}
\newcommand{\wtvu}{\textsc{w2v2-U2.0}\xspace}
\newcommand{\semantic}{\textsc{SSP}\xspace}
\newcommand{\semanticbase}{\textsc{SSP-base}\xspace}
\newcommand{\semanticbaser}{\textsc{SSP+R}\xspace}
\newcommand{\semanticbasera}{\textsc{SSP+RA}\xspace}
\newcommand{\semanticbaseap}{\textsc{SSP+AP}\xspace}
\newcommand{\semanticbest}{\textsc{SSP-tune}\xspace}
\newcommand{\semanticbestw}{\textsc{w2v2l14 + \semanticbest}\xspace}
\newcommand{\semanticbesth}{\textsc{Hubert + \semanticbest}\xspace}

\newcommand{\fsc}{\textsc{FSC}\xspace}
\newcommand{\slurp}{\textsc{SLURP}\xspace}
\newcommand{\slue}{\textsc{SLUE}\xspace}
\newcommand{\sqa}{\textsc{NMSQA}\xspace}
\newcommand{\squad}{\textsc{SQuAD-v1.1}\xspace}

\newcommand{\pipeline}{\textsc{Pipeline}\xspace}
\newcommand{\bert}{\textsc{BERT}\xspace}
\newcommand{\roberta}{\textsc{RoBERTa}\xspace}
\newcommand{\bart}{\textsc{BART}\xspace}
\newcommand{\longformer}{\textsc{Longformer}\xspace}
\newcommand{\dual}{\textsc{Dual}\xspace}

\newcommand{\lugosch}{\textsc{Lugosch}\xspace}

\newcommand{\cti}{\textsc{CTI}\xspace}

\begin{document}
\maketitle
\begin{abstract}
% \derek{TODO - Citations]
Recent studies find existing self-supervised speech encoders contain primarily acoustic rather than semantic information. As a result, pipelined supervised automatic speech recognition (ASR) to large language model (LLM) systems 
%outperform self-supervised speech embeddings 
achieve state-of-the-art results on semantic spoken language tasks by utilizing rich semantic representations from the LLM. These systems come at the cost of labeled audio transcriptions, which is expensive and time-consuming to obtain. We propose a task-agnostic unsupervised way of incorporating semantic information from LLMs into self-supervised speech encoders without labeled audio transcriptions. By introducing semantics, we improve existing speech encoder spoken language understanding performance by over 10\% on intent classification, with modest gains in named entity resolution and slot filling, and spoken question answering FF1 score by over 2\%. Our unsupervised approach achieves similar performance as supervised methods trained on over 100 hours of labeled audio transcripts, demonstrating the feasibility of unsupervised semantic augmentations to existing speech encoders. %effectiveness of self-supervised semantic speech encoders.
\end{abstract}

\section{Introduction}
% Self-supervised speech encoders compute universal speech representations from unlabelled audio data and achieve competitive results on spoken language understanding (SLU), where intent, named entities, and slot values are predicted from an input utterance, and spoken question answering (SQA), where a segment of an audio passage is identified as the solution to an audio question.
Realizing virtual AI assistants, artificial intelligence that can understand and respond to spoken language, is a north star for many natural and spoken language processing researchers. These systems typically follow an encoder-decoder architecture, where the encoder represents input audio as high-dimensional embeddings and the decoder converts said embeddings to downstream task outputs.
% \annie{I think you can move the intro of the downstream tasks to the beginning of the Introduction section, and the Abstract will be a concise summary of a) what's the problem, b) what's our proposal to solve this problem, c) what's the result}
% Thus, creating powerful task-agnostic semantically-rich speech encoders is an important step towards realizing this goal. 
To benchmark virtual AI assistants, researchers use semantic speech tasks, such as spoken language understanding (SLU) tasks, where intent, named entities, and slot values are predicted from an input utterance, and spoken question answering (SQA), where start and end frames of an audio passage are identified from an audio question.

%virtual AI assistants. 
% \annie{you may want to give a short intro of virtual AI assistants and why having task-agnostic self-supervised speech encoder which understands language semantics is important to virtual assistant. This will also help readers understand why " maintaining a separate large specialized model for each task is not computationally efficient".} 
Recent semantic speech benchmarks show systems with self-supervised transformer-based speech encoders achieve state-of-the-art performance against traditional filter bank approaches~\cite{yang2021superb, bastianelli2020slurp, shon2022slue}. A particularly notable setup is the universal representation setup~\cite{yang2021superb}, where a shared self-supervised speech encoder is pretrained upstream once and frozen for all downstream tasks, then a lightweight decoder is fine-tuned on each downstream task. This setup is appealing for building virtual AI assistants as maintaining a separate large specialized model for each task is not computationally efficient. Besides the universal representation setup, self-supervised speech encoders are also used by state-of-the-art specialized SLU systems to convert speech into high-dimensional representations~\cite{chung2020splat,kim2021st, qian2021speech, rao2021mean, agrawal2022tie, seo2022integration}. Hence, augmentations that universally boost performance on speech encoders trained with self-supervised learning (SSL) is of great interest to the spoken and natural language processing communities~\cite{schneider2019wav2vec, baevski2020wav2vec, hsu2021hubert, baevski2021unsupervised, liu2022towards}.

Recently, the speech community found SSL speech encoders capture primarily acoustic, not semantic, information~\cite{pasad2021layer}. \annie{this paragraph is very SLU specific. If we don't have SQA result, this is good. If we have SQA, it's considered as a very different task (you also see if from your experiments), it's better to start with discussing how people incorporate speech model and LM for semantic tasks in general, then go into the task-specific techniques that people use for SLU, SQA in particular.} \derek{For arxiv version, I think better to just include SLU, SQA dual model training should be done by Friday, bust still questions about fair setup. Update: SQA results out. Is this still a concern?} Thus, researchers propose end-to-end systems~\cite{chung2020splat, kim2021st, qian2021speech, le2022deliberation, seo2022integration, lin2022dual} that introduce semantic information through large language models (LLMs), such as \roberta~\cite{liu2019roberta} or \bart~~\cite{lewis2019bart}, which are pretrained to capture language semantics~\cite{clark2019does}, for SLU and SQA. An effective paradigm for combining SSL speech encoders with LLMs is the pipeline approach~\cite{bastianelli2020slurp}, which uses a bridge module to convert speech encoder embeddings into LLM subword tokens. The bridge module then pipes speech encoder output as input to the LLM~\cite{rao2021mean, lin2022dual, seo2022integration}. State-of-the-art solutions follow a 2-stage training process, where (1) audio to text transcription pairs are used to train the bridge module for automatic speech recognition (ASR), then (2) the speech encoder, bridge, and LLM modules are aligned by training the whole system end-to-end for SLU~\cite{lugosch2019speech, rao2021mean, lin2022dual, seo2022integration}. Unlike SSL speech encoders, this approach relies on ASR data, which is expensive to collect and maintain. Furthermore, existing pipelined models  are not task-agnostic, as they are trained for a single task or dataset. Thus, these models do not fit under the universal representation framework.

The multilingual speech translation community recently proposed effective unsupervised ASR models (ASR-U)~\cite{liu2020towards,baevski2021unsupervised, liu2022towards}. The state-of-the-art ASR-U model uses generative adversarial networks (GANs)~\cite{goodfellow2020generative} to learn to output text transcription from input audio recordings~\cite{liu2022towards}. Current ASR-U models output phoneme level transcriptions, though use various methods, such as Weighted Finite State Transducers (WFST) and self-training, to decode phoneme logits into raw text~\cite{baevski2021unsupervised, liu2022towards}. Our work leverages ASR-U to improve existing SSL speech encoders by treating it as a bridge module to inject semantic information into audio representations.
%\annie{we may want to write out the relationship between ASR-U and our work.} \derek{updated.} 

We propose an augmentation that introduces \textbf{\textit{\underline{S}}}emantics into existing SSL \textbf{\textit{\underline{SP}}}eech encoders, \semantic, by using ASR-U~\cite{liu2022towards} to extract information from LLMs. We adopt the pipelined approach to obtain semantic embeddings, with a convolutional neural network (CNN) + WFST bridge module trained on the GAN objective. As ASR-U is inherently noisy, we found an attention residual connection~\cite{he2016deep, vaswani2017attention} between the output of the SLL speech encoder and LLM can bring consistent performance improvement. We also propose an efficient way to align the LLM with the speech encoder through adapter~\cite{houlsby2019parameter} modules. Our final model achieves impressive performance gains on 3 SLU tasks and SQA across 4 datasets, under the universal representation setup. 

Our work serves as a valuable step to incorporating language semantics into self-supervised speech encoders. While there have been previous attempts using ASR-U to augment existing speech encoders with phoneme-level language models~\cite{feng2022superb,meng2022compressing}, subword-level LLMs contain much more pertinent and measurable semantic information~\cite{clark2019does}. Other works in spoken question answering rely on clustering to assign audio frames to frequent subword tokens, but this is both requires heavy finetuning on the downstream task and is not task-agnostic~\cite{lin2022dual}.
To the best of our knowledge, we are the first to propose a task-agnostic SSL speech encoder which directly interfaces with subword-based LLMs. We summarize our contributions below:

\begin{itemize}
  % \item We quantify the performance gap of SSL speech encoders that have or have not been injected with semantic information.
  \item We show how to effectively use ASR-U to interface SSL speech encoders with LLMs.
  \item We illustrate the importance of retaining both semantic and acoustic information with attention residual connections.
  \item We demonstrate the effectiveness of adapters in aligning speech and text encoders.
\end{itemize}
% \annie{1) we may want to cite jiatong's work and mention specifically what's the difference between ours and theirs, as their main claim is also "introducing language semantics to speech", 2) can we summarize our contribution into bullets so that it's easier for the readers to grasp?}
% Motivate from 
% pretrained language model encode semantics applied to speech angle?
% self-supervised speech model to encode semantics angle?
% semantic vs acoustic information?
\section{Related Works}

\subsection{Self-Supervised Speech Encoders}
SSL speech encoders are trained to learn and reconstruct pooled clustered representations of input audio from the original audio. The intuition for this objective comes from linguistics, where speech can be broken down into phoneme groups, where different chunks of input audio represent different phoneme groups. %SSL speech encoders intuitively tries to learn and reconstruct these groups.
%Self-supervised speech encoders rely on intuition that segments of input audio can be quantized into clustered representations. For example, spoken language is typically composed of sequences of phonemes, where each phoneme comes from a segment of input audio. 
\wtvo~\cite{schneider2019wav2vec} trains a convolutional neural network model to reconstruct the quantized cluster representations. \wtv~\cite{baevski2020wav2vec} uses transformers and a discrete codebook quantization module. \hubert~\cite{hsu2021hubert} improves \wtv by disentangling the clustering and SSL objectives and using a \bert-style encoder~\cite{devlin2018bert}. Benchmarks show that SSL speech encoders are the most effective method for solving multiple downstream tasks with minimal fine-tuning~\cite{yang2021superb}. A recent analytical work finds SSL speech encoders successfully encode acoustic information, but lack semantic information~\cite{pasad2021layer}. %found in large language models, such as \bert~\cite{devlin}, \roberta~\cite{liu2019roberta}, and \bart~\cite{lewis2019bart}. 

% Candidates for universal speech encoders broadly fall under traditional filterbank-based approaches and modern self-supervised transformer-based models, where the model uses unlabeled speech data to learn task-agnostic speech representations. Recent works have shown the latter achieves significantly better results than the former~\cite{superb} as optimizing self-supervised objectives can discover more acoustically rich features than handcrafted filter banks. Of the self-supervised methods, the most consistent encoders include Wav2Vec2 and Hubert~\cite{superb}. Wav2Vec2 ... Hubert ... However, recent studies show while such methods can capture acoustically rich representations, they fail to capture semantic information[Derek: Prof. Livescu's talk was on unpublished work, maybe refer to DUAL in this case?].

\subsection{Large Language Models}
In contrast to speech encoders, pretrained LLMs are shown to capture rich semantic information~\cite{clark2019does}. These methods optimize variants of the masked language modeling (MLM) objective to train a large transformer model. \bert~\cite{devlin2018bert} uses MLM to learn a transformer encoder. \roberta~\cite{liu2019roberta} introduces dynamic masking and a larger text corpus. \bart~\cite{lewis2019bart} supports generative modeling and adds a denoising objective, making it less susceptible to noisy text inputs. \longformer~\cite{beltagy2020longformer} is pretrained for long documents by increasing the document length limit during pretraining. LLMs have been successfully integrated with speech models for specific semantic tasks~\cite{chung2020splat, kim2021st, qian2021speech, le2022deliberation, seo2022integration, lin2022dual}, but not under the universal representation framework. % As pretrained large language models take sub-word tokens as input, to apply them on spoken language tasks, a connector layer must be proposed to convert input audio into subword tokens.

\subsection{Task-Specific Speech Models}
%Because SSL speech encoders fail to capture semantic information, s
State-of-the-art task-specific SLU systems adopt an end-to-end pipelined approach, which first converts input speech to embeddings with a SSL speech encoder, then converts embeddings to subword tokens via a bridge module, and finally feeds the subword tokens to a downstream model. One bottleneck is the bridge module. \lugosch~\cite{lugosch2019speech} trains a LSTM bridge module to convert audio features into phonemes then text. \cti's~\cite{seo2022integration} bridge module uses ASR logits to compute a weighted average of token embeddings. In addition to improving the bridge module, other works attempt to also distill LLM embeddings into speech representations~\cite{chung2020splat, cha2021speak, kim2021st, agrawal2022tie}. For optimizing targeted metrics, researchers have also experimented with reinforcement learning~\cite{rao2021mean}. While combinations of these methods achieve state-of-the-art performance, they require complicated training procedures with labeled ASR data, perform heavyweight training on the downstream dataset, and is not task-agnostic.

% The recent trend with state-of-the-art SLU or SQA [Derek: introduce SLU and SQA acronyms in intro] model are end-to-end pipelined models which take speech as input, converts it into subword tokens via a connector, and feeds the subword tokens into a large language model decoder. While this approach achieves impressive performance, it commonly relies on complicated training procedures that either do not generalize across tasks or requires fine-tuning a significant number of model parameters. Lugosch ... CoraJung ... Alexa ... CTI ... DUAL ... While none of these methods propose a task-agnostic speech encoder, they further emphasis the need for incorporating large language models with self-supervised speech encoders.

\subsection{Unsupervised ASR}
Recent work show the viability of unsupervised speech recognition. \wtvuo~\cite{baevski2021unsupervised} accomplished this by running Principal Component Analysis (PCA), k-means clustering, and mean pooling to convert \wtv~\cite{baevski2020wav2vec} features into phoneme-granularity features, then trains a GAN model to output phoneme text from the post-processed model~\cite{baevski2021unsupervised}. The state-of-the-art method
%\annie{of what constraints/setup} 
for phoneme-level unsupervised ASR is \wtvu~\cite{liu2022towards} which directly trains a CNN to output phonemes from \wtv features and uses a reconstruction loss to tie the input audio with corresponding generated text. Both methods use WFSTs to decode the phonemes into raw text. While there have been preliminary attempts~\cite{feng2022superb,meng2022compressing} to use \wtvu with phoneme language models\footnote{\url{https://huggingface.co/voidful/phoneme_byt5}}, we are the first to combine it with semantically-rich subword-based LLMs.

\subsection{Adapters}

Adapters are intermediary layers added to a large pretrained encoder. Adapter weights are learned during fine-tuning while the rest of the pretrained model is frozen. Adapters serve the dual purpose of efficient fine-tuning and preventing overfitting. First used by computer vision researchers~\cite{rebuffi2017learning}, adapters now enjoy much success in the natural language processing community by efficiently tuning LLMs~\cite{houlsby2019parameter}. In particular, the multilingual speech translation community found that adapters can effectively align SSL speech encoders and LLMs for spoken translation tasks~\cite{li2020multilingual,le2021lightweight}.

% \derek{Should I include a subsection on adapters?} \annie{yes as long as page limit allows} \derek{TODO: add subsection on adapters}
\section{Proposed Method} \label{sec:model}

% \begin{figure}
% \centering
% \includegraphics[width=0.12\textwidth]
% \label{fig:semantic-base}
% \end{figure}

% \begin{figure}
% \centering
% \includegraphics[width=0.12\textwidth]
% \label{fig:semantic-best}
% \end{figure}

\begin{figure}[h]
\centering
\includegraphics[width=0.25\textwidth]{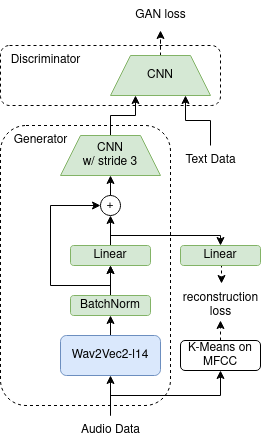}
\caption{Outline of \wtvu training procedure. The CNN module generates phoneme logits for the input audio. The bridge connector is trained on the GAN objective with reconstruction loss and regularization. During inference, the generator and linear layer used for reconstruction are discarded.}
\label{fig:w2vu}
\end{figure}

\begin{figure}
\centering
\begin{subfigure}{.25\textwidth}
  \centering
  \includegraphics[width=0.8\linewidth]{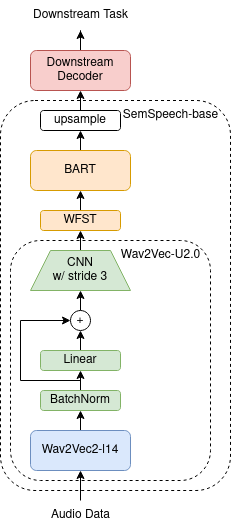}
  \caption{\semanticbase}
  \label{fig:semantic-base}
\end{subfigure}%
\begin{subfigure}{.25\textwidth}
  \centering
  \includegraphics[width=0.8\linewidth]{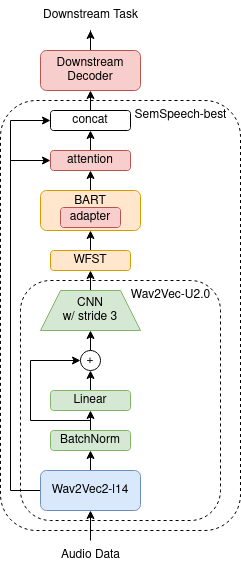}
  \caption{\semanticbest}
  \label{fig:semantic-best}
\end{subfigure}
\caption{Depiction of the \semantic. The blue component is the speech encoder, \wtvl, trained with SSL. The green component is the bridge module trained with a GAN objective. The orange component is the LLM, \bart, pretrained on a text corpus. The red components are trained on the downstream task.}
\label{fig:test}
\end{figure}

We propose to introduce semantics into SSL speech encoders by using ASR-U to interface with LLMs. 
%\wtv speech encoder with semantic information from \bart. \
% \annie{can we be more general here? BART is one LLM} 
Section~\ref{ssec:bridge} describes how to use ASR-U to link a speech encoder with a LLM. Section~\ref{ssec:residual} describes how to combine both acoustic and semantic information and deal with ASR transcriptions errors. Finally, Section~\ref{ssec:adapter} describes how to align LLMs with the speech encoder for downstream tasks.
% \derek{Should I add a section for Problem Setting?} \annie{yeah we should}

\subsection{Problem Setting}
Following the universal representation framework~\cite{yang2021superb}, our model consists of a heavy-weight speech encoder, $\mathcal{E}:\mathcal{X}\rightarrow\mathcal{Z}$, mapping input audio, $X \in \mathcal{X}$, to embeddings, $Z \in \mathcal{Z}$, and a light-weight task decoder, $\mathcal{D}: \mathcal{Z} \rightarrow \mathcal{Y}$, mapping embeddings to a probability distribution over the downstream task, $Y \in \mathcal{Y}$. The speech encoder, $\mathcal{E}$, is pretrained once, then shared on all downstream tasks. The task decoder, $\mathcal{D}$, is fine-tuned on each downstream task. During fine-tuning, the majority of model weights are frozen. This ensures we can efficiently store and deploy our model.

During pretraining, the speech encoder is trained on speech data, $X \in \mathcal{X}$, and unlabeled text data, $T_u \in \mathcal{T}_u$. During finetuning, the model is trained on labelled pairs from a downstream dataset, $(X,Y) \in \mathcal{X}\times\mathcal{Y}$. Notice, at no point are we given access to labelled ASR data, $(X,T_u) \in \mathcal{X}\times\mathcal{T}_u$, as it is costly to collect for different downstream datasets or languages.

\subsection{Unsupervised Semantic Representation as a Bridge} \label{ssec:bridge}
To incorporate semantic information into SSL speech encoders~\cite{baevski2020wav2vec, hsu2021hubert}, $\mathcal{E}_{acoustic}:\mathcal{X}\rightarrow\mathcal{Z}$, we wish to leverage subword-based LLMs~\cite{devlin2018bert, liu2019roberta, lewis2019bart, beltagy2020longformer}, $\mathcal{M}_{semantic}: \mathcal{S} \rightarrow \mathcal{Z}$, that capture language semantics. The major challenge to using LLMs in conjunction with speech encoders is the mismatch of input spaces. speech encoders take raw audio as input, $X \in \mathcal{X}$. LLMs take subword tokens as input, $S\in \mathcal{S}$. \semantic uses a bridge module~\cite{seo2022integration} to convert the speech encoder embedding output into subword tokens which can then be input into the LLM in a pipelined approach, $\mathcal{B}:\mathcal{Z}\rightarrow\mathcal{S}$.

As our upstream encoder is not trained with supervised ASR data, we use GANs~\cite{goodfellow2020generative} to train the bridge module, $\mathcal{B}$. Following \wtvu~\cite{liu2022towards}, we opt to generate phoneme sequences, $P\in\mathcal{P}$, conditioned on input audio. We do not directly predict subword-level transcriptions using our generator, because subword barriers are not easily deduceable from purely acoustic speech embeddings, requiring the model to implicitly learn phoneme-to-subword mappings. We find learning this information end-to-end is not easily achievable. Instead, we convert phoneme logits into subword tokens using a Weighted Finite State Transducer (WFST), $\mathcal{W}:\mathcal{P}\rightarrow\mathcal{S}$, which is fed known phoneme-to-subword mappings. We choose a convolutional neural network (CNN) as our phoneme generator, $\mathcal{G}:\mathcal{Z}\rightarrow\mathcal{P}$, and a separate CNN as our phoneme sequence discriminator, $\mathcal{C}:\mathcal{P}\rightarrow[0,1]$.

We adopt the same unsupervised training scheme as \wtvu~\cite{liu2022towards}. Specifically, we train the generator, $\mathcal{G}$, on GAN loss, $\mathcal{L}_{gan}$, a gradient penalty term, $\mathcal{L}_{gp}$, for better convergence, a smoothness penalty term, $\mathcal{L}_{sp}$, to encourage consecutive speech segments to generate the same phonemes, a phoneme diversity term, $\mathcal{L}_{pd}$, to diverse phoneme usage in output transcripts by maximizing entropy, and a self-supervised reconstruction loss, $\mathcal{L}_{ss}$, to encourage the generated phonemes to match the input audio. The reconstruction term uses a separate linear head to classify each speech embedding into 1 of 64 clusters, $\zeta_t$, obtained from running k-means on the Mel-frequency cepstral coefficient (MFCC) features of the input audio~\cite{hsu2021hubert, liu2022towards}. The final GAN training objective is summarized in Equation~\ref{eq:gan}. The training procedure for the bridge module is outlined in Figure~\ref{fig:w2vu}.

% \begin{equation}
\begin{gather}
\mathcal{L} =  \mathop{\mathrm{min}}_{\mathcal{G}} \mathop{\mathrm{max}}_{\mathcal{C}} [ \mathcal{L}_{gan}+ \lambda \mathcal{L}_{gp} + \gamma \mathcal{L}_{sp} + \eta \mathcal{L}_{pd} + \delta \mathcal{L}_{ss} ] \nonumber\\[1ex]
% + \lambda L_{gp}
\mathcal{L}_{gan} = \mathop{\mathbb{E}}_{T_u} [log\mathcal{C}(T_u)] + \mathop{\mathbb{E}}_X [log(1-\mathcal{C}(\mathcal{G}(X))] \nonumber\\
\mathcal{L}_{gp} =  \mathop{\mathbb{E}}_{\substack{X,T_u \\ \mu \sim U(0,1)}} [(|| \nabla \mathcal{C}(\mu \mathcal{G}(X) + (1-\mu ) T_u || - 1)^2] \nonumber\\
\mathcal{L}_{sp} = \sum_{(p_t,p_{t+1}) \in \mathcal{G}(X)} ||p_t - p_{t+1} ||^2 \nonumber\\
\mathcal{L}_{pd} = \frac{1}{|B|} \sum_{S\in B} -H_\mathcal{G}(\mathcal{G}(S)) \nonumber\\
\mathcal{L}_{ss} = -\sum_{t} logP_{\mathcal{G}}(\zeta_t|X).
\label{eq:gan}
\end{gather}
% \end{equation}

The vanilla version of our final model is composed of (1) SSL speech encoder, $\mathcal{E}_{acoustic}:\mathcal{X}\rightarrow\mathcal{Z}$ pretrained on unlabelled audio data, (2) a CNN+WFST bridge module, $\mathcal{B} = \mathcal{W}\circ\mathcal{G}:\mathcal{Z}\rightarrow\mathcal{S}$, trained on unlabelled text and audio data, and (3) a LLM, $\mathcal{M}_{semantic}:\mathcal{S}\rightarrow\mathcal{Z}$, pretrained on unlabelled text data. The input is fed to the SSL speech encoder and the output is read from the LLM. We also add an upsampling layer, $\mathcal{U}: \mathcal{Z}\rightarrow\mathcal{Z}$ to make the sequence length of the LLM output match the speech encoder output to better measure the \semantic's performance against just the speech encoder.

We choose the 14th layer of the \wtv~\cite{baevski2020wav2vec} as our speech encoder. The 14th layer is hypothesized to be better than the last layer as the last layers overfit the self-supervised training objective rather than learn informative acoustic representation. Previous work show if the speech encoder is frozen and only downstream decoder weights are fine-tuned for the ASR task, layer 14 performs much better than other layers~\cite{fan2020exploring, baevski2021unsupervised, pasad2021layer}. We choose \bart~\cite{lewis2019bart} as our LLM, as it is trained to denoise noisy input subword tokens, and we expect the bridge module to introduce some noise. We call this version of our model \semanticbase. A depiction can be found in Figure~\ref{fig:semantic-base}.

% As shown in previous works, large language models can deliver semantically rich embeddings that self-supervised speech encoders cannot.
% We wish to leverage large language models requiring an interface converting speech to text (ASR problem).
% We do not wish to use any ASR data because of problem statement.
% Use wav2vec-U and LLM.

\subsection{Combining Semantics and Acoustics with Residual Attention} \label{ssec:residual}
We hypothesize certain tasks may require more acoustic information than others. For instance, to perform named entity recognition (NER), the downstream model must implicitly transcribe named entities from input speech, but ASR relies primarily on acoustic, not semantic, information. Since the pipelined model may suffer from transcription errors introduced by ASR-U, naively using the pipelined approach introduces a bottleneck in acoustic information at the bridge module. To alleviate this issue, we propose adding a residual connection~\cite{he2016deep} from the output semantic embeddings to the original speech embedding.

% \begin{gather}
% H_t = [H_{at} || \mathrm{MHA}(H_{at}, H_{st}, H_{st})] \nonumber\\
% \mathrm{MHA}(Q,K,V) = [\alpha_0, \alpha_1, ..., \alpha_h] W_0 \nonumber\\
% \alpha_i = \frac{\mathrm{Softmax}((QW_{i}^0) (KW_{i}^1)^T)}{\sqrt{d_k}} (VW_{i}^2)
% \label{eq:resatt}
% \end{gather}

This can be done in 2 ways: (1) upsampling semantic embeddings and concatenating with speech embeddings, $Z = [Z_{\mathcal{E}} || \mathcal{U}(Z_{\mathcal{M}})]$, or (2) using multihead attention~\cite{vaswani2017attention} to merge the 2 embeddings,
%described by Equation~\ref{eq:resatt}
$Z = [Z_{\mathcal{E}} || \mathrm{MHA}(Z_{\mathcal{E}}, Z_{\mathcal{M}}, Z_{\mathcal{M}})]$, where $Z_{\mathcal{E}}\in\mathcal{Z}$ is the output of the \wtvl~\cite{baevski2020wav2vec} and $Z_{\mathcal{M}}\in\mathcal{Z}$ is the output of \bart~\cite{lewis2019bart}. The former is a simpler but more naive method. The latter is more effective as the attention layers are able to learn the alignment between speech and semantic embeddings. Notice, although the latter introduces slightly more learnable parameters finetuned on the downstream task, we find the number of new parameters to be inconsequential compared to the size of the lightweight downstream decoder, while providing significant performance improvement. % We call the former method \semanticres.

\subsection{Aligning Pretrained Text Model with Adapters} \label{ssec:adapter}
Inspired by works from speech translation~\cite{li2020multilingual,le2021lightweight}, we hypothesize that the LLM can easily be adapted for speech tasks through the use of adapters. We adopt the general recipe for adapters, where an adapter~\cite{houlsby2019parameter}, composed of a LayerNorm and 2-layer ReLU multilayer perceptron stack, is added to the end of each feed forward layer in the LLM and finetuned on the downstream task. Like residual attention, We find this small addition introduces a unsubstantial number of new parameters while providing a potentially substantial performance boost. We call the model using both residual attention and adapters \semanticbest, and outline it in Figure~\ref{fig:semantic-best}. \derek{should I add a figure here?} \annie{isn't it 2b?}\annie{shall we have another name for 2b, e.g. semspeech-enhanced, or something like that?}\derek{oh I mean do we need another figure illustrating adapters?}

% We hypothesize using purely semantic or acoustic representations is not enough for speech tasks.
\section{Experiments}

% \begin{figure}
% \centering
% \includegraphics[width=0.9\textwidth]{figures/tsne_w2v2-l14.png}
% \label{fig:tsne-ours}
% \end{figure}

% \begin{figure}
% \centering
% \includegraphics[width=0.9\textwidth]{figures/tsne_semspeech_best.png}
% \label{fig:tsne-baseline}
% \end{figure}

\begin{center}
\begin{table}
\begin{tabular}[width=0.5\textwidth]{|c | c | c |} 
 \hline
 Dataset & \# of Utterances & \# of Hours \\ [0.5ex] 
 \hline\hline
 \fsc-train & 23,132 & 14.7 \\ 
 \fsc-dev & 3,118  & 1.9 \\
 \fsc-test & 3,793 & 2.4 \\
 \hline
 \slurp-train & 50,628 & 40.2 \\
 \slurp-dev & 8,690 & 6.9 \\
 \slurp-test & 13,078 & 10.3 \\
 \hline
 \slue-train & 5,000 & 14.5 \\
 \slue-dev & 1,753 & 5.0 \\
 \hline
\end{tabular}
\caption{Dataset statistics for \fsc, \slurp, and \slue. Note, for \slue, only the train and dev splits are publicly available, thus we evaluate on the dev set.}
\label{tab:dataset-statistics}
\end{table}
\end{center}

\begin{table*}
\begin{center}
\begin{tabular}[width=0.5\textwidth]{|c | c c c c |} 
 \hline
 Model & \fsc-IC (Acc) & \slurp-IC (Acc) & \slurp-SF (F1) & \slue-NER (F1) \\ [0.5ex] 
 \hline
 \hline
 \wtv & 95.28\%  & 39.77\% & 36.48\% & 46.10\% \\
 % \hline
 % \hline
 \wtvl & 95.60\% & 49.97\% & 62.43\% & 78.77\% \\
 \hubert & 98.76\% & 58.11\% & 66.97\% & \bf{82.88\%} \\
 \hline
 \lugosch & 98.80\% & - & - & - \\
 \hline
 \semanticbase & 94.83\% & 55.28\% & 61.59\% & 79.62\% \\
 \semanticbestw & 98.71\% & 63.64\% & 64.48\% & 80.10\% \\
 % \semanticbaseh & 59.28\% & 66.30\% & 77.69\% \\
 \semanticbesth & \bf{99.44\%} & \bf{64.33\%} & \bf{68.82\%} & 82.02\% \\
 \hline
\end{tabular}
\caption{Experimental Results on \fsc, \slurp, and \slue datasets. We group the models by SSL encoders and their semantically-enriched counterparts. Note, the inclusion of semantic information consistently improves downstream performance significantly for both \wtvl and \hubert. For \slue-NER, the primary information type needed is acoustic. Hence, while \semanticbest helps \wtvl's less powerful acoustic embeddings, \hubert and \semanticbesth performs similarly.}
\label{tab:slurpslue} 
\end{center}
\end{table*}

% \begin{table}
% \begin{center}
% \begin{tabular}[width=0.5\textwidth]{|c | c |} 
%  \hline
%  Model & \fsc-IC (Acc) \\
%  \hline
%  \hline
%  % \wtvo & 84.92\% \\
%  % wtvb & 92.35 \\
%  \wtv & 95.28\% \\
%  % \hline
%  % hubert-base & 98.34 \\
%  % \hubert & \bf{98.76\%} \\
%  % \hline
%  \wtvl & 95.60\%\\
%  \hubert & 98.76\% \\
%  \hline
%  \lugosch & 98.80\% \\
%  \hline
%  \semanticbase & 94.83\% \\
%  \semanticbest & 98.71\% \\
%  % \semanticbaseh & 98.99\% \\
%  \semanticbesth & \bf{99.44\%} \\
%  \hline
%  % hubert-base & 98.34 \\
% \end{tabular}
% \caption{Experimental Results on \fsc. We group the models by different semantically-enriched SSL encoders, and a task-specific supervised encoder. Notice, inclusion of semantic information improves downstream performance significantly.}
% \label{tab:fsc}
% \end{center}
% \end{table} 

\subsection{Dataset}
To show the effectiveness of incorporating semantics into speech encoders, we benchmark \semantic against SSL speech encoders on various SLU tasks~\cite{yang2021superb}: intent classification (IC), slot filling (SF), and named entity recognition (NER). The goal of IC is to classify the intent of an input audio snippet. The goal of SF is to extract certain attributes of a given intent from an audio snippet. The goal of NER is to identify named entities in an audio snippet. We use 3 benchmark datasets: Fluent Speech Commands (\fsc)~\cite{lugosch2019speech}, Spoken Language Understanding Resource Package (\slurp)~\cite{bastianelli2020slurp}, and Spoken Language Understanding Evaluation (\slue)~\cite{shon2022slue}, which covers a wide variety of speakers, microphones, and settings.

To show the effectiveness of ASR-U as a bridge, we benchmark \semantic against a state-of-the-art unsupervised SQA method, \dual~\cite{lin2022dual}, on a recent benchmark SQA dataset, Natural Multi-speaker Spoken Question Answering (\sqa)~\cite{lin2022dual}, whose train set is generated synthetically, but test set is produced by human speakers. The goal of SQA is to find the start and end frames of a spoken answer in a spoken passage given a spoken question.

\subsubsection{\fsc}
The \fsc dataset~\cite{lugosch2019speech} is an IC dataset for a smart home virtual assistant. The input is a single audio file containing spoken English commands and the output class is the intent of the spoken command. The data was obtained through crowd-sourcing from 97 native and non-native English speakers. In total, there are 31 intents. The number of utterances and hours of each split can be found in the Table~\ref{tab:dataset-statistics}.

\subsubsection{\slurp}
The \slurp dataset~\cite{bastianelli2020slurp} is an IC and SF dataset for an in-home personal robot assistant. The input is a single audio file containing spoken English commands and the output is the scenerio, action, and entities. In total, there are 18 different scenarios, 46 different actions (IC), and 56 different entities (SF). The data was collected from 177 native and non-native English speaking Amazon Mechanical Turk workers. SLURP is a more challenging dataset than FSC as SLURP use both headsets and microphones with various microphones placements. The number of utterances and hours of each split can be found in Table~\ref{tab:dataset-statistics}

\subsubsection{\slue}
The \slue dataset~\cite{shon2022slue} is a NER dataset using European Parliament event recordings. The input is a single audio file containing spoken English passages and the output are the named entities. There are in total 7 categories that were based on the OntoNotes Release 5.0~\cite{hovy2006ontonotes} entity labels. The dataset was collected from the official European Parliament website. SLUE demonstrates the effectiveness of our model on more specialized audio data. The number of utterances and hours of each split can be found in the Table~\ref{tab:dataset-statistics}.

\subsection{\sqa}
The \sqa dataset~\cite{lin2022dual} is a SQA dataset generated from a standard text question answering dataset, \squad~\footnote{A question answering dataset using Wikipedia articles}, using Amazon Polly Text-to-Speech~\footnote{https://aws.amazon.com/tw/polly} for the train and dev split, and 60 human speakers for the test set. \sqa demonstrates the effectiveness of different unsupervised semantic speech embedding approaches. \sqa contains 297.18 hours, 37.61 hours, and 2.67 hours of train, dev, and test split audio respectively. We follow \dual by evaluating on Frame-level F1 score (FF1) and Audio Overlapping Score (AOS). 

\begin{table*}
\begin{center}
\begin{tabular}[width=0.5\textwidth]{|c | c c c c |} 
 \hline
 Augmentation & \fsc-IC (Acc) & \slurp-IC (Acc) & \slurp-SF (F1) & \slue-NER (F1) \\ [0.5ex] 
 \hline
 \hline
 None & 95.60\% & 49.97\% & 62.43\% & 78.77\% \\
 % \hline
 \semanticbase & 94.83\% & 55.28\% & 61.59\% & 79.62\% \\
 \semanticbaser & 97.55\% & 59.59\% & 63.37\% & 79.66\% \\
 \semanticbasera & \bf{98.97\%} & 62.39\% & 64.21\% & 80.04\% \\
 \semanticbaseap & 96.07\% & 60.28\% & 63.85\% & 79.97\% \\
 % \hline
 \semanticbest & 98.71\% & \bf{63.64\%} & \bf{64.48\%} & \bf{80.10\%} \\
 \hline
\end{tabular}
\caption{Ablation studies on the residual attention and adapters when applied to \wtvl. Notice, both components are important to best utilize acoustic and semantic information. While \semanticbasera introduces slightly more parameters than \semanticbaser, it provides tangible performance improvement by better aligning the acoustic and semantic embeddings.}
\label{tab:ablation}
\end{center}
\end{table*}

\begin{table}
\begin{center}
\begin{tabular}[width=0.5\textwidth]{|c | c |} 
 \hline
 Model Component & \% of Parameters \\ % (\%) \\
 \hline
 \hline
 % \wtvl & 23,132 \\ 
 % bridge & 3,118 \\
 % \bart & 100M \\
 \semanticbase & 90.40\% \\ %456,811,008 \\ %(90.40\%) \\
 \hline
 downstream decoder & 8.69\% \\ %43,951,172 \\ %(8.70\%) \\
 residual attention & 0.73\% \\ %3,674,112 \\ %(0.73\%) \\
 \bart adapters & 0.18\% \\ %894,528 \\ %(0.18\%) \\
 \hline
\end{tabular}
\caption{Comparing the parameter count of different components of \semanticbest. In total, there are 505.3 million parameters. Notice, the downstream decoder is much more lightweight than the upstream model. Residual attention and adapters also introduce minimal parameter overhead to finetuning.}
\label{tab:params}
\end{center}
\end{table}
% DUAL
% \subsubsection{Spoken Language Understanding}
% TODO

% \subsubsection{Spoken Question Answering}
% TODO

\subsection{Baselines}
To show \semantic improves existing speech encoders, we compare it against existing state-of-the art SSL speech encoders. We compare against \wtv~\cite{baevski2020wav2vec}, which uses SSL to train a transformer model, \wtvl~\cite{baevski2020wav2vec}, which uses the 14th layer as the output rather than the last layer, and \hubert, which improves \wtv's SSL objective to train a \bert-style model. 

As mentioned in Section~\ref{sec:model}, we compare SSL speech encoders against 2 versions of our model, \semanticbase and \semanticbest. The former uses the pipelined approach to connect \wtvl with \bart~\cite{lewis2019bart} with no additional modifications. The latter introduces an attention residual connection and learnable adapters to combine acoustics and semantics together and align the LLM with the speech encoder respectively. We either connect the residual connection to the output of \wtvl, yielding \semanticbestw, or to the output of \hubert, yielding \semanticbesth.

%To better measure the performance gains from incorporating semantics, we also compare against \wtv~\cite{baevski2020wav2vec} and \hubert~\cite{hsu2021hubert} on all datasets. \annie{here we are comparing with hubert, but semspeech is not using hubert. in this case, do we also need results for hubert-semspeech?} \derek{I think the reasoning here is the performance improvement from semspeech is comparable to the improvement from wav2vec2 to hubert. The con is that people might start to question if we need semspeech if we have hubert. Arguably +semspeech is more generalizable than / is parallel to hubert. Given introducing hubert introduces these issues, should we not report it until we have hubert-semspeech?} \wtv~\cite{baevski2020wav2vec} uses the last layer rather than the 14th layer used by \wtvl. \hubert~\cite{hsu2021hubert} improves \wtv by using offline clustering and BERT embeddings for SSL, and serves as a benchmark to compare relative performance gains from semantics versus better SSL training. 

On the \fsc dataset, we compare against \lugosch~\cite{lugosch2019speech}, a supervised LSTM-based model specialized for the IC task. It is not task-agnostic and uses labeled ASR data. Unlike the other \wtv-based or \hubert-based baselines, \lugosch does not use SSL or transformers. 

To show \semantic improves existing unsupervised sematic speech methods, we compare it against the existing state-of-the-art unsupervised SQA method, \dual~\cite{lin2022dual}, which uses clustering on \hubert to obtain frame-level tokens. These tokens are then used to train a \longformer model with linear classification head. To fairly compare with \dual, which fine-tunes the LLM, we use \semanticbase but perform lightweight LLM adapter tuning. We call this resulting model, \semanticbaseap. To produces frame-level predictions, we remove the upsampling layer from \semanticbaseap . We choose \wtvl as our speech model and \bart as our LLM, as it is robust to ASR errors. We also show a \pipeline model, which trains a \wtv model on ASR data and a \longformer LLM on text-only question answering data. Notice, since evaluation is based on the frame-level, SSL speech encoders are not a baseline since they operate at the audio level.

Ideally, only minimal parameter updates are needed for the downstream task, thus we adopt various lightweight decoders from the speech processing universal performance benchmark (SUPERB)~\cite{yang2021superb}. For IC, the decoder is a sum pooling layer followed by a multilayer perceptron classifier and trained with cross entropy loss. For the SF and NER tasks, the decoder is recursive neural network (RNN) model that transcribes input audio into text. The decoder identifies named entities or slot values by surrounding them with named special tokens. For instance, to identify ``fox" as the subject in the sentence, ``the quick brown fox jumped over the lazy dog.", the decoder would predict: ``the quick brown [SUBJECT-start] fox [SUBJECT-end] jumped over the lazy dog." The RNN decoder is trained with connectionist temporal classification loss. For SQA, we adopt the same decoder as \dual~\cite{lin2022dual}, which is a linear layer classifying each subword embedding as the start or end or neither of an answer span.
% hubert

\begin{table*}
\begin{center}
\begin{tabular}[width=0.9\textwidth]{|c | c | c c | c c c | c c |} 
 \hline
 \multirow{2}{*}{Bridge Module} & \multirow{2}{*}{ASR data} & \multicolumn{2}{c|}{\fsc} & \multicolumn{3}{c|}{\slurp} & \multicolumn{2}{c|}{\slue} \\
 \cline{3-9}
 & & WER & IC Acc & WER & IC Acc & SF F1 & WER & NER F1 \\
 \hline
 \hline
 % \hline
 \wtvasr & 960h & \bf{9.19\%} & \bf{99.34\%} & \bf{45.83\%} & \bf{66.18\%} & \bf{65.62\%} & \bf{15.51\%} & \bf{80.58\%} \\
 \wtvasr & 100h & 11.89\% & 99.10\% & 53.22\% & 63.20\% & 63.87\% & 17.70\% & 79.67\% \\
 \wtvasr & 10h & 59.06\% & 98.50\% & 74.77\% & 59.91\% & 63.42\% & 53.00\% & 79.76\% \\
 \semanticbest & nothing & 21.28\% & 98.71\% & 51.51\% & 63.64\% & 64.48\% & 31.22\% & 80.10\% \\
 \hline
\end{tabular}
\caption{Analysis on WER of \semantic's bridge module. All models adopt the same speech encoder, LLM, residual attention, and adapter components as \semanticbestw, but convert speech embeddings into subword tokens in different ways. \wtvasr finetunes \wtv with an ASR head using letter-based CTC with varying amounts of ASR data. As seen in this table, ASR errors correlate with downstream performance. Hence, more accurate ASR-U models or components that alleviate ASR errors, such as residual attention, would greatly benefit semantic speech encoders.} %\bart, residual attention, and adapters are able to keep downstream performance high, despite the bridge module producing occasional transcription errors. \derek{maybe can't claim this for FSC; should we compare with supervised approaches?}}
\label{tab:sup}
\end{center}
\end{table*}

\begin{figure}
\centering
\begin{subfigure}{.5\textwidth}
  \centering
  \includegraphics[width=1.0\linewidth]{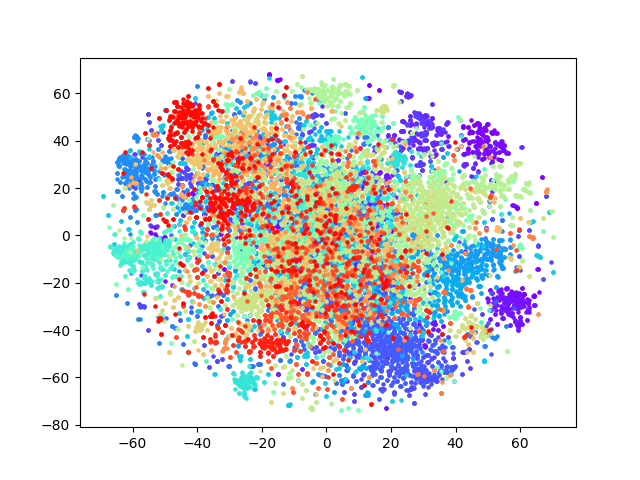}
  \caption{t-SNE embeddings using \wtvl encoder}
  \label{fig:tsne-baseline}
\end{subfigure}%
\\
\begin{subfigure}{.5\textwidth}
  \centering
  \includegraphics[width=1.0\linewidth]{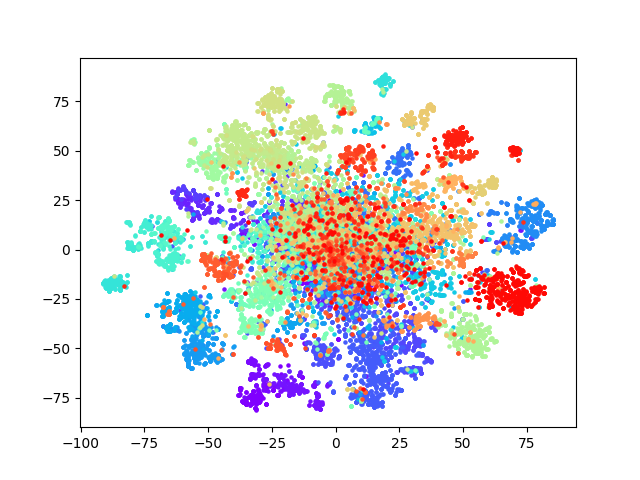}
  \caption{t-SNE embeddings using \semanticbest encoder}
  \label{fig:tsne-ours}
\end{subfigure}
\caption{t-SNE visualizations of pooled audio snippet embeddings for \slurp-IC. Each point corresponds with one embedding. The color denotes the ground truth class of the corresponding audio snippet. Subfigure~\ref{fig:tsne-baseline} and ~\ref{fig:tsne-ours} shows \semanticbest is better at differentiating intents by incorporating semantic information.}
\label{fig:tsne}
\end{figure}

\subsection{Spoken Language Understanding}

As seen in Table~\ref{tab:slurpslue}, \semantic significantly improves the performance of both \wtvl and \hubert. The performance gap between a \wtvl model with and without semantics is similar to the gap between \wtvl and the task-specific \lugosch model. Note, \hubert, which uses \bert-style SSL, also benefits from \semantic, confirming that semantics are a significant bottleneck in existing SSL speech encoders. \semanticbesth performs the best overall as it fully augments the best existing SSL speech encoder, \hubert.

The relative performance gain is more for IC than SF and NER, since SF and NER requires the speech encoder to transcribe identified entities, which includes performing ASR. For such ASR-adjacent tasks, the residual connection in \semanticbest is crucial, as it prevents acoustic information loss across the bridge connector's information bottleneck. We see large performance gains in IC since the semantics introduced by \bart greatly improves the effectiveness of the speech encoder. Combining both acoustic and semantic information, as done by \semanticbest models, provides the most consistent performance improvement, since the skip connection can learn which type of information is more needed.

% As seen in Table~\ref{tab:fsc}, \semantic significantly improves the performance of both \wtvl and \hubert. The performance gap between a \wtvl model with and without semantics is similar to the gap between \wtvl and the task-specific \lugosch model. Note, \hubert, which uses \bert-style SSL, also benefits from \semantic, confirming that semantics are a significant bottleneck in existing SSL speech encoders. \semanticbesth performs the best overall as it fully augments the best existing SSL speech encoder, \hubert.

% Table~\ref{tab:slurpslue} shows this trend holds on a variety of tasks and datasets. The relative performance gain is more for IC than SF and NER, since SF and NER requires the speech encoder to transcribe identified entities, which includes performing ASR. For such ASR-adjacent tasks, the residual connection in \semanticbest is crucial, as it prevents acoustic information loss across the bridge connector's information bottleneck. We see large performance gains in IC since the semantics introduced by \bart greatly improves the effectiveness of the speech encoder. Combining both acoustic and semantic information, as done by \semanticbest models, provides the most consistent performance improvement, since the skip connection can learn which type of information is more needed.

\subsection{Spoken Question Answering}
As seen in Table~\ref{tab:sqa}, \semantic outperforms recent unsupervised clustering-based approaches, because ASR-U tokens are better aligned with LLMs than \hubert clusters. Unlike \dual, \semantic better utilizes pretrained LLMs, since the LLM input space is the same as \semantic's output space. Furthermore, \semantic does not require careful hyperparameter tuning of \hubert cluster counts.

\begin{table}
\begin{center}
\begin{tabular}[width=0.5\textwidth]{|c | c c |} 
 \hline
\multirow{2}{*}{Model} & \multicolumn{2}{c|}{\sqa} \\ % (\%) \\
 \cline{2-3}
& FF1 & AOS \\
 \hline
 \hline
 % \wtvl & 23,132 \\ 
 % bridge & 3,118 \\
 % \bart & 100M \\
 \dual-64 & 39.0\% & 33.0\% \\ %43,951,172 \\ %(8.70\%) \\
 \dual-128 & 55.9\% & 49.1\% \\ %43,951,172 \\ %(8.70\%) \\
 \dual-512 & 17.3\% & 12.5\% \\ %43,951,172 \\ %(8.70\%) \\
 \semanticbaseap & 57.20\% & 46.44\% \\ %456,811,008 \\ %(90.40\%) \\ % 57.19783 46.44346
 \semanticbaseap\textsuperscript{\textdagger} & \bf{58.18\%} & \bf{52.09\%} \\ %456,811,008 \\ %(90.40\%) \\
 \hline
 \pipeline\textsuperscript{\textdagger} & \bf{64.2\%} & \bf{57.1\%} \\ %3,674,112 \\ %(0.73\%) \\
 \hline
\end{tabular}
\caption{Comparing unsupervised SQA models to supervised \pipeline model. \textdagger\xspace denotes the model uses a LLM that was finetuned on the \squad text-only QA dataset. We compare the baseline, \dual, with 3 different number of clusters choices, to \semanticbaseap trained with either unlabeled audio or text.}
\label{tab:sqa}
\end{center}
\end{table}

\subsection{Ablation Study}
% Using adapters on the language model is critical for aligning the embeddings on downstream tasks.
% Merging semantic and acoustic embeddings is also critical for performance boost.
To more carefully analyze the affect of residual attention and adapters in \semanticbest, we run experiments on all SLU datasets with and without each component. We denote these two design choices as +RA and +AP respectively. As seen in Table~\ref{tab:ablation}, both components provide ample performance improvement over the vanilla \semanticbase model. We also confirm the amount of additional parameters for slot filling introduced by these learnable components are unsubstantial compared to the existing number of parameters in the downstream decoder in Table~\ref{tab:params}. Notice, the downstream task finetunes less than 10\% of total model parameters. \semanticbest require less than 1\% more parameters than \semanticbase during downstream fine-tuning.

We also try the naive residual connection approach described in Section~\ref{ssec:residual} by upsampling and concatenating semantic embeddings to the speech embeddings. We call this approach \semanticbaser. This method is less effective than \semanticbasera as it is not trained to align the speech and semantic embeddings together, but still improves \semanticbase, further validating our hypothesis that merging acoustic and semantic information is beneficial.

% Our approach to unsupervised semantic embeddings outperform previous SOTA specially designed for SQA

\subsection{Comparison with Supervised ASR Methods}
% How good are the transcriptions generated by the U-ASR method for downstream tasks? As good as supervised methods trained on 100 hours of labeled training data.
To quantify the effect of transcription errors introduced by the bridge module, we compute the word error rate (WER) of the bridge connector in \semanticbest, and compare it against standard \wtv supervised ASR models~\cite{baevski2020wav2vec} trained on 10 minutes, 100 hours, and 960 hours of labeled ASR data. Table~\ref{tab:sup} confirms that less noisy transcripts, transcripts with lower WER, correlates with better downstream performance. The unsupervised model, which uses 960 hours of unlabelled data, can reach similar WER as a supervised model trained on 100 hours of labelled data, indicating the effectiveness of the bridge module. On \slurp and \slue, the relative drop in WER ($>20\%$) is substantially more than the relative drop in downstream performance ($<5\%$), verifying \semanticbest's tolerance to noisy transcriptions. The robustness to ASR errors come from our choice of LLM, \bart, which is trained to handle noisy inputs, residual connection to acoustic embeddings, and LLM alignment with adapters.
\annie{can we have another error analysis sub-section after you have the error analysis results?}\derek{depends on what to include; need to rerun ASR experiments, as rn the letter-based vs LM based decoding schemes aren't easily comparable; Suyoun also gave good suggestion of separating WER of correct and erroneous predictions to show better ASR-U could improve our method.}
% We analyze the noise introduced by the bridge component by measuring the word error rate of the output transcriptions. While the bridge does produce nontrivial ASR errors, we find that the performance cost of these errors are minimal by comparing against a supervised ASR model trained on varying amounts of supervised data. As seen in Table~\ref{TODO}, the ASR model trained on 960 hours of super ised Librispeech data can achieve relatively low word error rate, yet the downstream performance ois similar to Semantic (best), despite the fact that it has much higher WER.  This is because the BART model can tolerate ASR errors since it was trained to denoise input text, as well as the residual attention and adapter components augmenting the semantic embeddings with acoustic representations.

\subsection{Representation Visualization}
% TODO: PLOT T-SNE EMBEDDINGS FOR INTENT CLASSIFICATION
To better see the impact of including semantic representations, we visualize the pooled audio snippet embedding for intent classification on \slurp-IC using t-distributed stochastic neighbor embedding (t-SNE)~\cite{van2008visualizing}. We denote the ground truth label of each audio snippet by the color of its pooled embedding. As seen in Figure~\ref{fig:tsne}, the clusters produced by semantic embeddings are more spread out and better separated than those produced by just acoustic speech embeddings, indicating that \semantic introduces new semantic information that existing speech encoders lack.

\section{Conclusion}
We presented a compelling case for introducing semantics into self-supervised speech encoders and an effective method of doing so. Our approach boosts the performance of existing speech encoders on multiple different spoken language understanding tasks and datasets. We also provide reasoning for what tasks may benefit more or less from incorporating semantics. Furthermore, our approach is task agnostic and can augment any existing model using  SSL speech encoders. With \semanticbest, we show how better utilize semantic embeddings by merging acoustic and semantic information and effectively aligning LLMs to the speech encoder on downstream tasks with minimal parameter overhead. As it can generalize to many downstream tasks, \semantic provides an important step towards actualizing virtual AI assistants.
\printbibliography

\end{document}